\documentclass{article}

\makeatletter
\def\input@path{{./}{icml/}}
\makeatother

\PassOptionsToPackage{table}{xcolor}
\usepackage{microtype}
\usepackage{graphicx}
\graphicspath{{icml/}}
\usepackage{subcaption}
\usepackage{booktabs} %

\usepackage{hyperref}
\usepackage[most]{tcolorbox}
\usepackage{xcolor}
\usepackage{tabularx}
\usepackage{arydshln}
\usepackage[utf8]{inputenc}
\usepackage[T1]{fontenc}
\usepackage{sourcesanspro}


\usepackage[preprint]{icml2026}

\usepackage{amsmath}
\usepackage{amssymb}
\usepackage{mathtools}
\usepackage{amsthm}
\usepackage{enumitem}
\usepackage[capitalize,noabbrev]{cleveref}

\theoremstyle{plain}

\theoremstyle{definition}

\theoremstyle{remark}

\definecolor{lightblue}{RGB}{235,245,255}
\definecolor{lightgreen}{RGB}{240,255,240}
\definecolor{lightorange}{RGB}{255,240,230}
\definecolor{lightpurple}{RGB}{250,240,255}

\usepackage[textsize=tiny]{todonotes}

\icmltitlerunning{Verbalized Action Masking for Controllable Exploration in RL Post-Training}

\begin{document}

\twocolumn[
  \icmltitle{VAM: Verbalized Action Masking for Controllable Exploration \\ in RL Post-Training -- A Chess Case Study}

  \begin{icmlauthorlist}
    \icmlauthor{Zhicheng Zhang}{cmu}
    \icmlauthor{Ziyan Wang}{kcl}
    \icmlauthor{Yali Du}{kcl}
    \icmlauthor{Fei Fang}{cmu}
  \end{icmlauthorlist}

  \icmlaffiliation{cmu}{Carnegie Mellon University}
  \icmlaffiliation{kcl}{King's College London}
  \icmlcorrespondingauthor{Zhicheng Zhang}{zhichen3@cs.cmu.edu}

  \icmlkeywords{Machine Learning, ICML}

  \vskip 0.3in
]

\printAffiliationsAndNotice{}  %

\begin{abstract}
  Exploration remains a key bottleneck for reinforcement learning (RL) post-training of large language models
  (LLMs), where sparse feedback and large action spaces can lead to premature collapse into repetitive behaviors.
  We propose \textit{Verbalized Action Masking (VAM)}, which verbalizes an action mask in the prompt and enforces that
  the model outputs an action from the masked set.
  Building on this interface, we introduce iterative action-space pruning: if the target action is not sampled, we
  remove valid sampled actions from the mask and resample under the reduced candidate set, repeating until the target
  is sampled or a fixed budget is exhausted.
  We study VAM in chess and evaluate it under two training regimes: an engine-play regime that generates states
  via play against an engine opponent and a fixed-dataset regime that trains from a fixed dataset of positions
  with verifier scores.
  Across held-out chess puzzles and full-game play measured by average centipawn loss (ACPL), VAM improves learning
  efficiency and final performance over strong baselines, highlighting verbalized masking as a practical mechanism for
  controllable exploration in LLM RL post-training.
\end{abstract}

\section{Introduction}
\label{sec:introduction}
\begin{figure*}[htbp]
    \centering
    \includegraphics[width=0.95\linewidth]{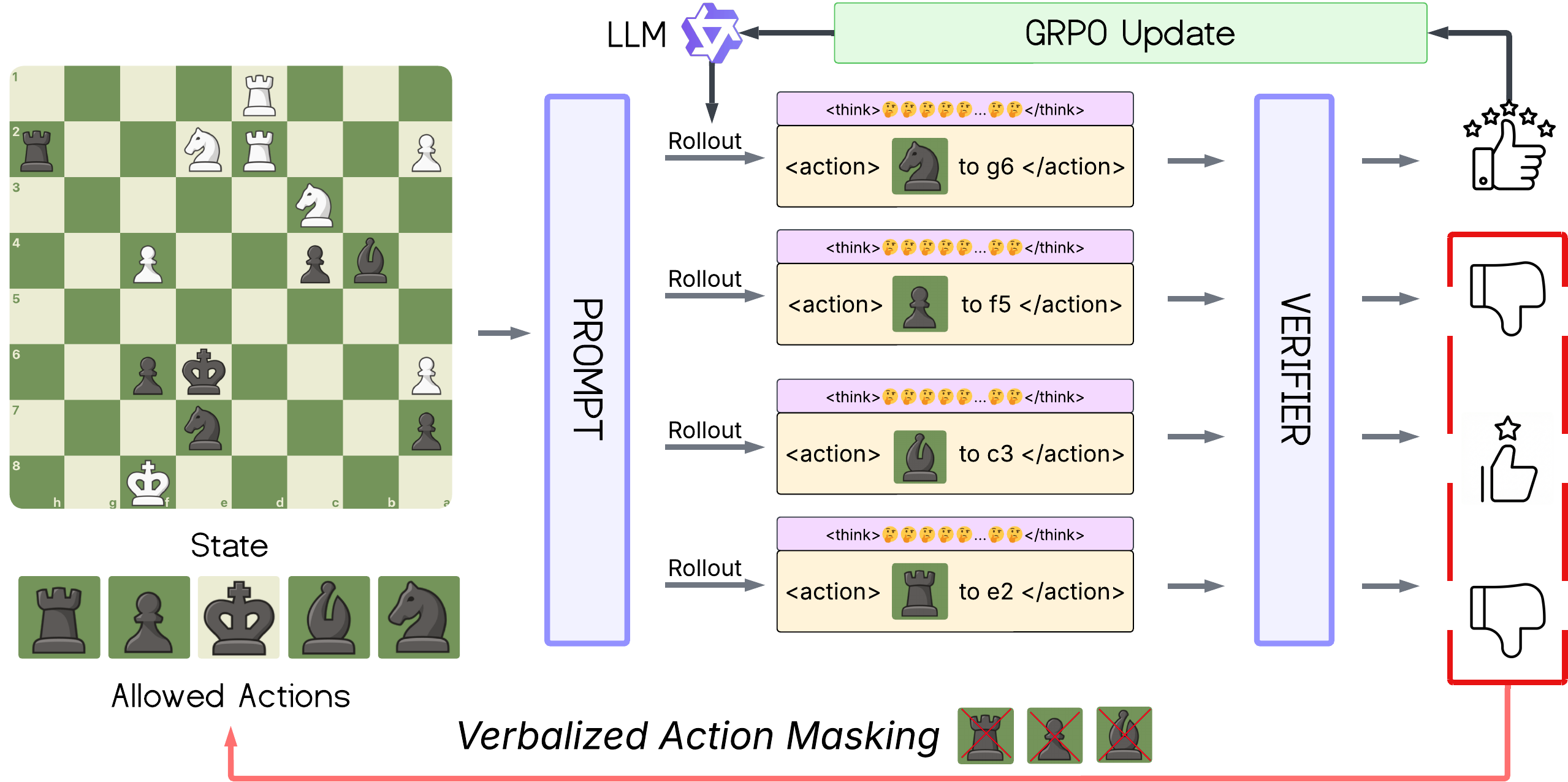}
    \caption{\textbf{Overview of Verbalized Action Masking (VAM) with iterative action-space pruning.}
Given a state and a provided set of allowed actions, we prompt the LLM and sample rollout groups for Group
Relative Policy Optimization (GRPO).
Each sampled output is parsed into an action and then evaluated by a verifier to produce reward feedback.
If the target condition is not met in the current round (for example, the target action is not sampled), we mask the
distinct valid sampled actions out of the allowed action set and resample under the updated prompt.
This iterative pruning procedure increases within-state action coverage and yields more informative grouped on-policy
updates.}
    \label{framework}
\end{figure*}
Reinforcement learning with verifiable rewards (RLVR) offers a scalable recipe for post-training large language models
(LLMs)~\citep{cobbe2021training,lightman2023let,openai2024openaio1card,gemmateam2025gemma3technicalreport}, where the model is trained against automatically checkable signals,
such as exact correctness or strict output-format compliance.
This ``verifier-first'' framing enables fully automated, on-policy RL loops and has driven strong improvements on
reasoning-heavy tasks in recent work~\citep{guo2025deepseek}.

RLVR, however, does not remove RL's central difficulty: exploration.
When the reward is sparse and the effective action space is large, on-policy updates stall if the policy does not
sample informative behaviors, even with deterministic verification.
In many LLM post-training settings, this failure mode is easy to miss because the ``environment'' is often just an
evaluator.
The learning problem is unchanged: the policy must discover high-reward actions (valid outputs under a strict interface)
in an immense space of possible strings.

Classical RL often explores by visiting new states through environment interaction.
In contrast, many RL post-training setups sample prompts from a fixed distribution and perform multiple rollouts per
prompt, for example with grouped on-policy objectives such as Group Relative Policy Optimization (GRPO)~\citep{guo2025deepseek}.
In this regime, the key exploration bottleneck is action selection within a single state.
Given the same prompt, does the model actually try different plausible actions, or does it repeatedly sample the same
high-probability output?
When a rollout group collapses to near-identical actions, group-relative updates provide little learning signal, and
additional samples largely waste compute.

We propose \textit{Verbalized Action Masking (VAM)} to make action-space exploration controllable.
The prompt explicitly lists an admissible action set, and a deterministic verifier checks that the model outputs a valid
action from this set.
Choosing an out-of-mask action receives a fixed penalty and terminates, which turns action validity into a verifiable
signal suitable for RLVR.
This is a prompt-level interface rather than logit-level masking; each rollout is sampled on-policy from the model conditioned
on the presented candidate set.
We then use this interface to enforce exploration via \emph{iterative action-space pruning}.
For a fixed state, we run multiple rounds of grouped sampling, remove already-sampled valid actions from the candidate
set, and resample under the pruned set until a target action is found or a fixed budget is exhausted.
This procedure converts repeated sampling into controlled coverage of the action space, rather than repetition of the
same few actions.

We study these ideas in chess, which provides an unusually clean testbed.
Each position admits a finite, enumerable action set, the legal moves, and legality checking is exact.
Moreover, strong engine evaluators can supply per-move value estimates that serve as dense, deterministic rewards.
At the same time, chess exposes practical failures of language-model policies.
Prior work, including \citet{hwang2025can} and LLM CHESS~\citep{kolasani2025llmchessbenchmarkingreasoning},
has found that providing legal-move lists and enforcing structured outputs are crucial for learning and evaluation, and
that interface choices can dominate performance.
We instantiate VAM for chess move selection, where the model must output a single move from a provided candidate list in
a strict format, and we evaluate both one-step move selection on held-out puzzles and full-game play
against an engine opponent using average centipawn loss (ACPL).

In summary, our contributions are the following:
\begin{enumerate}
	    \item We formalize action masking as an MDP augmentation that turns action validity into a verifiable reward and
	    termination signal;
		    \item We propose VAM with iterative action-space pruning, enabling controllable action-space exploration under grouped on-policy RL;
	    \item We present two practical training pipelines, fixed-dataset training and engine-play position generation, that share the same pruning mechanism;
	    \item We conduct extensive experiments in chess, demonstrating that VAM improves learning efficiency and final performance over strong baselines on both puzzle selection and full-game play.
\end{enumerate}

\section{Related Work}
\paragraph{RL with verifiable rewards (RLVR)}
Recent work has shown that reinforcement learning with automatically checkable outcome rewards can improve LLM
reasoning without relying on human preference labels.
This aligns with approaches in code generation that use execution feedback as a reward signal~\citep{le2022coderl,yang2023intercode}. DeepSeek-R1~\citep{guo2025deepseek} uses GRPO-style post-training with rule-based correctness signals and explicit output-format constraints,
demonstrating that verifiable rewards and structured generation can meaningfully shape model behavior.
Because our setting also admits deterministic verification (output parsing and legality checks in chess), we adopt a
similar ``verifier-first'' philosophy for reward design.
\citet{wen2025reinforcement} caution that commonly reported pass@$k$ improvements can mask whether reasoning
itself improved; they propose CoT-aware variants of pass@$k$ that require both a correct final answer and a correct
chain-of-thought.

\paragraph{Exploration in RL post-training}
RL post-training with verifiable rewards can suffer from exploration failures, where on-policy updates collapse diversity
even when an outcome verifier is available.
Recent work addresses this challenge with a variety of mechanisms, including outcome-level exploration incentives,
trial-and-error interaction for agentic environments, and partial supervision that scaffolds rare long-horizon solutions.
Outcome-based exploration~\cite{song2025outcome} adds bandit-inspired bonuses or within-batch repetition penalties over
final outcomes to encourage diverse solution attempts under GRPO-style training.
Exploration-Based Trajectory Optimization (ETO)~\citep{song2024trial} alternates between rolling out an LLM agent to collect failure
trajectories and updating the policy with trajectory-level preference optimization that contrasts failures against
higher-reward trajectories.
Complementary analyses study how RLVR can sharpen pass@1 while reducing pass@$k$ diversity, and propose algorithmic knobs
that preserve exploration, such as clipping variants for entropy control and reference-policy resets under
KL regularization~\citep{deng2025trial,park2025clip,liu2025prorl}.
Within this landscape, hint-based and off-policy guidance methods inject partial rationales, solution prefixes, or expert
traces during training to scaffold rare long-horizon solutions, including StepHint~\cite{zhang2025stephint},
AdaBack~\cite{amani2025rl}, privileged on-policy exploration (POPE)~\cite{qu2026pope}, and branched rollouts from
expert anchors (BREAD)~\cite{zhang2025bread}. Related approaches include off-policy guidance
(LUFFY)~\cite{yan2025learning} and self-improvement methods such as STaR~\cite{zelikman2022star} and
Self-Refine~\cite{madaan2023self}.
Our approach targets the same bottleneck, but rather than revealing rationales or shaping outcome rewards, we intervene
on the \emph{action space}: we present an explicit candidate set via VAM and enforce coverage by pruning already-sampled
actions across rounds.

\paragraph{Chess-playing LLMs}
Several recent works study whether LLMs can acquire chess competence through either supervised learning or RL~\cite{veeriah2025evaluating,feng2025generating}.
\citet{hwang2025can} perform GRPO post-training of LLMs on chess puzzles and report that providing explicit legal-move lists
is important for learning; their analysis suggests that performance plateaus and that limited chess-rule
understanding of chess rules is a bottleneck.
In contrast, ChessLLM~\citep{zhang2025complete} studies large-scale supervised training on complete games and reports
strong engine-based evaluations, highlighting the role of long-horizon data and evaluation in full-game settings.
ChessGPT and ChessCLIP~\citep{feng2023chessgpt} treat chess as a language modeling and alignment problem, pairing game records with natural
language annotations and proposing a broad evaluation suite spanning state tracking, value judgment, and tactical
policy probes.
Finally, LLM CHESS~\citep{kolasani2025llmchessbenchmarkingreasoning} benchmarks chess play through an agentic tool interface and finds that instruction-following and
tool-use failures, not only illegal moves, are a major source of breakdown, and that interface choices can materially
affect performance.
These findings motivate our focus on candidate-restricted move selection with strict output parsing and explicit
action masks.

\section{Preliminaries}
\label{sec:preliminaries}

\paragraph{Chess notation and legal moves}
We represent chess positions as Forsyth--Edwards Notation (FEN) strings and moves as Universal Chess Interface (UCI)
strings (e.g.,
\texttt{e2e4}, \texttt{g1f3}).
We also refer to Standard Algebraic Notation (SAN) for compact, human-readable moves (e.g., \texttt{e4}, \texttt{Nf3},
\texttt{O-O}) when discussing prior work.
SAN can also include annotations such as \texttt{+} (check) and \texttt{\#} (checkmate), which are convenient for
humans but can make tactical consequences explicit and thus act as an additional move-quality cue; we therefore use
strict UCI for model outputs and treat SAN only as auxiliary notation.
For a position $s$, let $\mathcal{A}(s)$ denote the set of legal moves.
In our experiments and evaluation, models are required to output a single move in strict UCI format, and malformed
outputs (or illegal moves) are treated as invalid.
To quantify move quality during full-game play, we use average centipawn loss (ACPL) as an engine-based metric.

\paragraph{Group Relative Policy Optimization (GRPO)}
\label{sec:prelim_grpo}
We post-train the policy using Group Relative Policy Optimization (GRPO)~\citep{guo2025deepseek}, an on-policy method
that estimates advantages from \emph{within-prompt groups} of sampled responses, avoiding a learned critic/value
function.
Given prompts $q \sim P(Q)$, GRPO samples a group of $G$ outputs $\{o_i\}_{i=1}^G \sim \pi_{\theta_{\text{old}}}(O\mid q)$,
computes scalar rewards $r_i$ and group-relative advantages
\begin{equation}
A_i \;=\; \frac{r_i - \mathrm{mean}(\{r_1,\dots,r_G\})}{\mathrm{std}(\{r_1,\dots,r_G\})},
\end{equation}
and optimizes the clipped objective
\begin{equation}
\begin{split}
J_{\text{GRPO}}(\theta)
&=
\mathbb{E}_{\substack{q \sim P(Q)\\ \{o_i\}_{i=1}^G \sim \pi_{\theta_{\text{old}}}(O\mid q)}}\!\Bigg[
\frac{1}{G}\sum_{i=1}^{G}\Big(
\min\!\big(\rho_i(\theta)A_i,\\
&\qquad \mathrm{clip}(\rho_i(\theta),1-\epsilon,1+\epsilon)A_i\big)
\\
&\qquad -\beta\,D_{\mathrm{KL}}(\pi_\theta \,\|\, \pi_{\mathrm{ref}})\Big)
\Bigg],
\end{split}
\end{equation}
where $\rho_i(\theta)=\pi_\theta(o_i\mid q)/\pi_{\theta_{\text{old}}}(o_i\mid q)$, $\epsilon>0$ is the PPO-style
clipping parameter, and $\beta\ge 0$ weights a KL penalty to a reference policy $\pi_{\mathrm{ref}}$ (computed per
sampled output).

\section{Methods}
\label{sec:methods}

Action-space exploration has been a central challenge in RL post-training with verifiable rewards.
In grouped on-policy objectives such as Group Relative Policy Optimization (GRPO)~\citep{shao2024deepseekmath}, each
prompt is paired with a group of sampled outputs, and learning depends on reward differences within the group.
When the per-state action space is large, the policy can collapse to a small subset of high-probability actions, yielding
low diversity and weak learning signals within each group.
We propose Verbalized Action Masking (VAM), which makes the admissible action set an explicit part of the RL
interface.
Specifically, we (i) formalize an action-masking MDP that assigns a fixed penalty when the policy selects an
out-of-mask action, (ii) implement this interface for LLM policies by verbalizing the mask in the prompt and enforcing
it with a deterministic parser and verifier, and (iii) improve exploration via iterative action-space pruning, which
repeatedly samples GRPO-sized rollout groups for the same base prompt while shrinking the mask to avoid resampling
previously produced valid actions.
VAM can be trained under two regimes that differ only in how states and verifier scores are obtained: a \emph{fixed-dataset}
regime that samples states from a fixed dataset with precomputed verifier scores, and an \emph{engine-play} regime that
generates states on-policy, for example by playing against an engine opponent and querying the verifier on encountered positions.
We first present VAM in a generic action-selection setting and then describe the chess instantiation used in our
experiments.

\subsection{Action-masking MDP}
Intuitively, an action mask specifies which actions are currently admissible.
If the agent selects an action outside this mask, the environment treats it as a constraint violation, assigns a fixed
penalty, and terminates the rollout.
This turns action validity into a deterministic verifier signal that can be used by an on-policy RL algorithm, and it
provides a clean handle for manipulating the available action space across repeated sampling.

Formally, consider a Markov decision process (MDP) $(\mathcal{S},\mathcal{A},P,R,\gamma)$ with states
$s\in\mathcal{S}$, actions $a\in\mathcal{A}$, transition kernel $P(s'\mid s,a)$, reward function $R(s,a)$, and discount
factor $\gamma\in[0,1)$.
Let $\mathbf{1}[\cdot]$ denote the indicator function.
We assume that at each state $s$ the environment can enumerate a finite admissible action set
$\mathcal{A}(s)\subseteq\mathcal{A}$.
An action mask is a subset $\mathcal{M}\subseteq\mathcal{A}(s)$ of actions that are currently allowed.
An \emph{action-masking MDP} augments the state with an explicit mask variable.
Specifically, define an augmented state space
\begin{equation}
\tilde{\mathcal{S}} \;=\; \{(s,\mathcal{M}) : s\in\mathcal{S},\; \mathcal{M}\subseteq \mathcal{A}(s)\},
\end{equation}
and an action-masking MDP $(\tilde{\mathcal{S}},\mathcal{A},\tilde{P},\tilde{R},\gamma)$.
The mask may depend on the underlying state and can be updated as a function of the chosen action; we write
$U(\mathcal{M}'\mid s,a,\mathcal{M})$ for a (possibly deterministic) mask-update kernel.
Then for in-mask actions the augmented transition factorizes as
\begin{equation}
\begin{split}
\tilde{P}\bigl((s',\mathcal{M}') \mid (s,\mathcal{M}), a\bigr)
&=
P(s'\mid s,a)\,U(\mathcal{M}'\mid s,a,\mathcal{M}).
\end{split}
\end{equation}
If $a\notin\mathcal{M}$, we treat the transition as terminal (equivalently, a transition to an absorbing terminal state),
so the next-state dynamics and mask update are not applied.

The augmented reward assigns the base reward for in-mask actions and a fixed penalty otherwise:
\begin{equation}
\begin{split}
\tilde{R}\bigl((s,\mathcal{M}), a\bigr)
&=
\mathbf{1}[a \in \mathcal{M}] \cdot R(s, a)
\\
&\quad+
\bigl(1 - \mathbf{1}[a \in \mathcal{M}]\bigr)\cdot (-\mathrm{Penalty}),
\end{split}
\end{equation}
where $\mathrm{Penalty}>0$ is a constant.
We terminate the rollout immediately when an out-of-mask action is chosen.
Equivalently, define a done indicator
\begin{equation}
d(s, a, \mathcal{M}) \;=\; \mathbf{1}[a \notin \mathcal{M}],
\end{equation}
and terminate whenever $d(s,a,\mathcal{M})=1$.
The penalty controls whether the mask is a soft constraint or a hard constraint.
When $\mathrm{Penalty}$ is finite, the objective discourages out-of-mask actions while still defining a reward value for
all actions.
In the limiting case $\mathrm{Penalty}\to\infty$, any out-of-mask action incurs an unbounded loss, and the optimal
policy assigns zero probability mass to actions outside $\mathcal{M}$, recovering a hard version of action masking.

\begin{algorithm}[t]
\caption{VAM with iterative action-space pruning}
\label{alg:vam_training}
\begin{algorithmic}[1]
\STATE \textbf{Procedure} \textsc{PruneAndSample}$(\pi, s, a^\star)$
\STATE \textit{Parameters:} group size $G$, max rounds $R_{\max}$, reward $\tilde{R}$
\STATE $\mathcal{M} \leftarrow \mathcal{A}(s),\; \mathcal{D}_s \leftarrow \emptyset$
\FOR{$r=1$ \TO $R_{\max}$}
    \STATE Sample $\{o_i\}_{i=1}^G \sim \pi(\cdot\mid s,\mathcal{M})$ and parse to $\{a_i\}_{i=1}^G$
    \STATE Compute $r_i \leftarrow \tilde{R}((s,\mathcal{M}),a_i)$ for $i=1,\dots,G$
    \STATE $\mathcal{D}_s \leftarrow \mathcal{D}_s \cup \{(\mathcal{M}, \{(o_i,r_i)\}_{i=1}^G)\}$
    \STATE $V \leftarrow \{a_i : a_i \in \mathcal{M}\}$
    \STATE \textbf{if} $a^\star \in V$ \textbf{then}
    \STATE \hspace{1em}\textbf{break}
    \STATE $\mathcal{M} \leftarrow \mathcal{M}\setminus V$
\ENDFOR
\STATE \textbf{return} $\mathcal{D}_s$
\STATE \textbf{Procedure} \textsc{TrainVAM}$(\pi, \mathcal{D}, T, B)$
\STATE \textit{Inputs:} policy $\pi$, dataset $\mathcal{D}$, total steps $T$, batch size $B$
\FOR{$t=1$ \TO $T$}
    \STATE Sample $\{s_j\}_{j=1}^B$ from $\mathcal{D}$ and set $\mathcal{B} \leftarrow \emptyset$
    \FOR{$j=1$ \TO $B$}
        \STATE Read $\mathcal{A}(s_j)$ and set $a^\star \in \arg\max_{a\in\mathcal{A}(s_j)} \mu(s_j,a)$
        \STATE $\mathcal{D}_{s_j} \leftarrow \textsc{PruneAndSample}(\pi, s_j, a^\star)$
        \STATE $\mathcal{B} \leftarrow \mathcal{B} \cup \mathcal{D}_{s_j}$
    \ENDFOR
    \STATE Update $\pi$ with GRPO on $\mathcal{B}$ (\S\ref{sec:prelim_grpo})
\ENDFOR
\STATE \textbf{return} $\pi$
\end{algorithmic}
\end{algorithm}

\subsection{Verbalized Action Masking (VAM)}
The action-masking MDP specifies how a mask should affect reward and termination, but it does not dictate how the mask
is represented to the policy.
In the LLM setting, we implement action masking by verbalizing the mask in the prompt.
Each prompt contains a description of the state $s$ and an explicit action mask $\mathcal{M}\subseteq \mathcal{A}(s)$.
The model then outputs a single action, and a deterministic parser and verifier map the output to a predicted action $\hat a$ and
check membership in $\mathcal{M}$.
If parsing fails or if $\hat a\notin\mathcal{M}$, we treat the outcome as an out-of-mask action, triggering termination and
the penalty as defined by the action-masking MDP.

\definecolor{promptBg}{HTML}{FFFFFF}
\definecolor{promptFrame}{HTML}{D8E0E6}
\definecolor{promptAccent}{HTML}{1B7A6F}
\definecolor{promptLabel}{HTML}{556579}
\definecolor{promptText}{HTML}{111827}
\definecolor{promptRowA}{HTML}{EAF5F3}
\definecolor{promptRowB}{HTML}{F2F4F7}
\definecolor{promptRowC}{HTML}{F8F1EA}
\newcommand{\promptfont}{\fontfamily{SourceSansPro-TLF}\selectfont}
\newcommand{\promptlabelfont}{\promptfont}
\newcommand{\promptboxsize}{\footnotesize}
\newcommand{\promptvaluefont}{\ttfamily\fontsize{9.0}{10.2}\selectfont}
\newcommand{\promptcode}[1]{{\color{promptText}#1}}
\newcommand{\promptcodecompact}[1]{{\color{promptText}#1}}
\newcommand{\promptformat}{{\promptcode{\shortstack[l]{<think>...</think>\\<action>...</action>}}}}
\newcommand{\promptlist}{{\promptcode{[act$_1$, \ldots, act$_k$]}}}
\newlength{\promptlabelwidthvam}
\newlength{\promptlabelwidthchess}
\newlength{\promptlabelwidthtmp}
\settowidth{\promptlabelwidthvam}{{\promptlabelfont\promptboxsize Format specification}}
\settowidth{\promptlabelwidthtmp}{{\promptlabelfont\promptboxsize State description}}
\ifdim\promptlabelwidthtmp>\promptlabelwidthvam
  \setlength{\promptlabelwidthvam}{\promptlabelwidthtmp}
\fi
\settowidth{\promptlabelwidthtmp}{{\promptlabelfont\promptboxsize Allowed action}}
\ifdim\promptlabelwidthtmp>\promptlabelwidthvam
  \setlength{\promptlabelwidthvam}{\promptlabelwidthtmp}
\fi
\addtolength{\promptlabelwidthvam}{0.05em}
\settowidth{\promptlabelwidthchess}{{\promptlabelfont\promptboxsize Original chess prompt}}
\settowidth{\promptlabelwidthtmp}{{\promptlabelfont\promptboxsize Legal moves (UCI)}}
\ifdim\promptlabelwidthtmp>\promptlabelwidthchess
  \setlength{\promptlabelwidthchess}{\promptlabelwidthtmp}
\fi
\settowidth{\promptlabelwidthtmp}{{\promptlabelfont\promptboxsize Allowed action}}
\ifdim\promptlabelwidthtmp>\promptlabelwidthchess
  \setlength{\promptlabelwidthchess}{\promptlabelwidthtmp}
\fi
\addtolength{\promptlabelwidthchess}{0.05em}
\newcolumntype{L}[1]{>{\raggedright\arraybackslash\color{promptLabel}\promptlabelfont\hyphenpenalty=10000\exhyphenpenalty=10000}m{#1}}
\newcommand{\promptstrut}{\rule[-0.55em]{0pt}{1.55em}}
\newcommand{\promptstruttight}{\rule[-0.4em]{0pt}{1.35em}}
\newcommand{\promptrow}[2]{\promptstrut #1 & \promptstrut {\promptvaluefont #2} \\}
\newcommand{\promptrowtight}[2]{\promptstruttight #1 & \promptstruttight {\promptvaluefont #2} \\}
\newtcolorbox{promptbox}[2][]{
    enhanced,
    colback=promptBg,
    colframe=promptFrame,
    arc=1.0mm,
    boxrule=0.35pt,
    boxsep=0mm,
    left=0.8mm,
    right=0.8mm,
    top=0.35mm,
    bottom=0mm,
    borderline west={1.0pt}{0pt}{promptAccent},
    fonttitle=\bfseries\promptfont\footnotesize,
    coltitle=white,
    attach boxed title to top left={xshift=1.5mm, yshift=-1.1mm},
    boxed title style={
        colback=promptAccent,
        colframe=promptAccent,
        arc=0.9mm,
        boxrule=0pt,
        top=0.35mm, bottom=0.35mm, left=0.95mm, right=0.95mm
    },
    title={#2},
    before upper=\setlength{\parindent}{0pt}\setlength{\parskip}{0pt}\promptboxsize\promptfont\color{promptText},
    #1
}
\noindent\begin{minipage}{\linewidth}
\begin{promptbox}{Verbalized Action Masking (VAM)}
\setlength{\tabcolsep}{1.6pt}
\setlength{\extrarowheight}{0pt}
\renewcommand{\arraystretch}{1.0}
\renewcommand{\tabularxcolumn}[1]{>{\raggedright\arraybackslash}m{#1}}
\begin{tabularx}{\linewidth}{@{}L{\promptlabelwidthvam}X@{}}
\rowcolor{promptRowA}\promptrow{State description}{\promptcode{\{task prompt\}}}
\rowcolor{promptRowB}\promptrow{Format specification}{\promptformat}
\rowcolor{promptRowC}\promptrowtight{Allowed action}{\promptlist}
\end{tabularx}
\end{promptbox}
\vspace{2pt}
\captionof{figure}{Prompt interface for VAM. Prompt inputs include the state description, format specification, and allowed action list; placeholders indicate where each appears.}
\end{minipage}

VAM preserves on-policy data collection.
The mask is part of the observation, and we sample from the current policy conditioned on the prompt, without applying
logit-level masking or post-hoc filtering.
As a result, each sampled group is on-policy for its corresponding mask-conditioned prompt, which makes the approach
compatible with standard grouped on-policy RL implementations.

\subsection{Iterative action-space pruning}
The action mask defines which actions are admissible, but it does not by itself ensure diverse exploration within a
single state.
To avoid repeatedly sampling the same few actions, we introduce an iterative pruning procedure that progressively
shrinks the mask across rounds of grouped sampling.

\paragraph{Verifier-based rewards and target actions}
\label{sec:methods_reward}
In RL post-training with verifiable rewards, a task reward is computed from the model output using an external verifier.
We assume that for each state $s$ and admissible action $a\in\mathcal{A}(s)$, the verifier provides a scalar score
$\mu(s,a)\in\mathbb{R}$.
This score may be binary (for example, an exact correctness check) or graded (for example, a test score or a heuristic
quality metric).
We use it to define the base reward $R(s,a)=\mu(s,a)$ for in-mask actions.
To make the interface verifiable, we also define a format reward over the raw output string $o$:
if $o$ does not match the required tag format (for example, \texttt{<action>...</action>} or \texttt{<uci\_move>...</uci\_move>}),
or if the parsed action $\hat a$ is not in $\mathcal{M}$, then $r_{\text{format}}(o,\mathcal{M})=-\mathrm{Penalty}$, and otherwise
$r_{\text{format}}(o,\mathcal{M})=0$.
This corresponds to treating unparsable outputs and out-of-mask actions as constraint violations, assigned a fixed penalty via $\tilde{R}$,
and terminated as in the action-masking MDP.

Given a state $s$ and a mask $\mathcal{M}\subseteq\mathcal{A}(s)$, we define the target action as
\begin{equation}
a^\star(s,\mathcal{M}) \in \arg\max_{a\in\mathcal{M}} \mu(s,a),
\end{equation}
using a deterministic tie-break when multiple actions share the maximum score.

For each base state $s$, the pruning procedure starts from an initial mask $\mathcal{M}_0\subseteq\mathcal{A}(s)$.
In the engine-play regime we use the full legal set $\mathcal{M}_0=\mathcal{A}(s)$, while in the fixed-dataset regime
$\mathcal{M}_0$ can be either the full legal set or a dataset-provided candidate subset.
We set the target to $a^\star=a^\star(s,\mathcal{M}_0)$.
Each round uses a new prompt that differs only in the masked action list, and samples a rollout group of $G$ candidates,
matching the GRPO group size (\S\ref{sec:prelim_grpo}).
If the target action is not present, we remove the sampled in-mask actions from the candidate set and resample, up to a
maximum number of rounds $R_{\max}$.
Each pruning round defines a separate mask-conditioned prompt for GRPO, so the procedure naturally produces a sequence of
GRPO groups per base state.
We apply GRPO to all collected groups without additional weighting across rounds.
Algorithm~\ref{alg:vam_training} summarizes the full VAM training loop, including iterative pruning, resampling, and the
GRPO update.
Algorithm~\ref{alg:online_play} shows an optional engine-play procedure for generating training states on-policy; when
engine-play position generation is not used, we assume a provided fixed dataset of states with verifier scores.

\begin{algorithm}[t]
\caption{Engine-play position generation (optional)}
\label{alg:online_play}
\begin{algorithmic}[1]
\STATE \textbf{Procedure} \textsc{CollectByPlay}$(\pi, E, \mathrm{Verifier}, B)$
\STATE \textit{Inputs:} environment $E$, target data size $B$, pool size $N$
\STATE Initialize or reuse $\mathcal{E}=\{e_1,\dots,e_N\}$ from $E$, and set $\mathcal{D}_{\text{play}}\leftarrow\emptyset$
\WHILE{$|\mathcal{D}_{\text{play}}| < B$}
    \STATE Retrieve current states $\{s_i\}_{i=1}^N$ from $\mathcal{E}$
    \FOR{$i=1$ \TO $N$}
        \STATE Read $\mathcal{A}(s_i)$ and query $\mathrm{Verifier}$ for $\mu(s_i, a)$ for all $a \in \mathcal{A}(s_i)$
        \STATE Append $(s_i, \mathcal{A}(s_i), \mu(s_i, \cdot))$ to $\mathcal{D}_{\text{play}}$
        \STATE \textbf{if} $|\mathcal{D}_{\text{play}}| \ge B$ \textbf{then}
        \STATE \hspace{1em}\textbf{break}
    \ENDFOR
    \STATE Step all instances in $\mathcal{E}$
\ENDWHILE
\STATE \textbf{return} $\mathcal{D}_{\text{play}}$
\end{algorithmic}
\end{algorithm}
\subsection{Chess implementation details}
\label{sec:methods_chess}
Our experiments instantiate VAM in chess move selection.

\paragraph{State, actions, and prompt interface}
We represent states $s$ as chess positions (FEN strings), and actions $a$ as moves in strict UCI format.
The admissible action set $\mathcal{A}(s)$ is the set of legal moves.
To implement VAM, each prompt includes an explicit candidate list corresponding to a mask
$\mathcal{M}\subseteq\mathcal{A}(s)$ (either the full legal set or a restricted subset), together with the ordered legal
move list.

\noindent\begin{minipage}{\linewidth}
\begin{promptbox}{Chess Prompt with Verbalized Mask}
\setlength{\tabcolsep}{1.6pt}
\setlength{\extrarowheight}{0pt}
\renewcommand{\arraystretch}{1.0}
\renewcommand{\tabularxcolumn}[1]{>{\raggedright\arraybackslash}m{#1}}
\begin{tabularx}{\linewidth}{@{}L{\promptlabelwidthchess}X@{}}
\rowcolor{promptRowA}\promptrow{Original chess prompt}{\promptcode{\{original chess prompt\}}}
\rowcolor{promptRowB}\promptrow{Current FEN string}{\promptcode{\{FEN\}}}
\rowcolor{promptRowB}\promptrow{Legal moves (UCI)}{\promptcode{\{UCI move list\}}}
\rowcolor{promptRowC}\promptrowtight{Allowed action}{\promptcode{\{allowed UCI move list\}}}
\end{tabularx}
\end{promptbox}
\vspace{2pt}
\captionof{figure}{Chess prompt with verbalized mask. Prompt inputs include the original prompt, FEN, legal moves, and allowed actions; placeholders indicate where each appears.}
\end{minipage}

The model must output a single move in a \texttt{\textless uci\_move\textgreater} tag.
A deterministic parser and verifier checks that the output is well-formed, that it is legal in the underlying position,
and that it lies in the provided mask.
Appendix~\ref{sec:appendix_experiments_parsing} specifies the exact parsing and validity rules.

\paragraph{Engine-based verifier scores}
We instantiate $\mu(s,a)$ using the Stockfish chess engine~\citep{stockfish}.
Given fixed analysis settings, Stockfish can provide different deterministic per-move evaluation signals, including
centipawn scores, win-draw-loss probabilities, and expected-score proxies.
Any chosen scalar signal can be used as the verifier score $\mu(s,a)$ to define rewards and targets.
In our experiments, training positions are paired with engine-derived per-move value maps (e.g., expected-score or
win-probability proxies) that serve as $\mu(s,a)$.

\paragraph{Fixed-dataset training}
In the fixed-dataset regime, training states $s$ are sampled from a fixed dataset, and each state comes with a legal-move list
and an engine-derived value map $\mu(s,a)$ over legal actions.
The initial mask $\mathcal{M}_0$ can either be the full legal set $\mathcal{A}(s)$ or a candidate subset provided by
the dataset.
Training follows the VAM loop in Algorithm~\ref{alg:vam_training}, which performs iterative pruning and applies a GRPO update over all collected mask-conditioned groups.

\paragraph{Engine-play position generation}
For engine-play training data, we generate a stream of chess positions by playing games against a fixed engine opponent.
Because the learning problem only requires a state, we record positions regardless of which side is to move.
For each recorded state $s$, we enumerate legal moves $\mathcal{A}(s)$, query the engine-derived verifier score
$\mu(s,a)$ over legal actions, and add the resulting state records to a training buffer.
Algorithm~\ref{alg:online_play} summarizes this optional engine-play position generation procedure; these states can then be
consumed by the same VAM training loop in Algorithm~\ref{alg:vam_training}.

\section{Experiments}
\label{sec:experiments}
We evaluate VAM in a text-only chess setting, in which the model must select a move from a provided list without
access to search or external tools.
We report (i) one-step move selection on the Searchless Chess puzzle positions~\citep{ruoss2024grandmaster} and
(ii) full-game play against a fixed engine opponent, measured by average centipawn loss (ACPL).

\subsection{Research questions}
\label{sec:experiments_rqs}
We organize our empirical study around three questions:
\begin{enumerate}[leftmargin=*, itemsep=0pt, topsep=2pt, parsep=0pt, partopsep=0pt]
    \item \textbf{RQ1:} Does VAM improve exploration compared to the baselines?
    \item \textbf{RQ2:} Does VAM improve full-game play compared to the baselines?
    \item \textbf{RQ3:} What are the relative strengths of engine-play position generation and a fixed dataset?
\end{enumerate}

\subsection{Experimental setup}
\label{sec:experiments_setup}
\paragraph{Models}
We run experiments with Qwen2.5-3B-Instruct and Qwen2.5-7B-Instruct~\citep{qwen2025qwen25technicalreport}.

\paragraph{Verbalized-masking interface}
Each decision prompt provides the position as a FEN string, the side to move, and an ordered list of legal moves in
strict lowercase UCI.
In the verbalized masking setting, we additionally provide a candidate subset,
\texttt{allowed\_moves}, and require the model to choose from this subset.
Models output a single move in strict lowercase UCI inside a \texttt{\textless uci\_move\textgreater} tag, and we evaluate with a
deterministic parser and legality checker.
We use the same prompt template as \citet{hwang2025can} (Appendix~\ref{sec:appendix_experiments_details}).
Appendix~\ref{sec:appendix_experiments_parsing} specifies the exact parsing and validity rules.

\paragraph{Training regimes and data}
We focus on post-training methods that learn from deterministic, engine-provided signals.
Each training position is paired with an engine-derived per-move value map $\mu(s,a)\in[0,1]$ over legal moves, which we
use to define reward signals (\S\ref{sec:methods_reward}).
We consider two sources of training positions: a fixed dataset aligned with Chess-R1~\citep{hwang2025can} containing
100{,}000 positions, and an on-policy stream of positions generated via engine-play against a fixed engine opponent.
These correspond to the fixed-dataset and engine-play regimes in \S\ref{sec:methods}.
In both cases, training follows the same VAM loop in Algorithm~\ref{alg:vam_training}, which applies GRPO to all
mask-conditioned groups produced by iterative pruning.
In the engine-play regime, training states are generated on-policy via play against an engine opponent using
Algorithm~\ref{alg:online_play}, while in the fixed-dataset regime we assume a provided dataset.
In RQ1 we use fixed-dataset training to isolate the effect of pruning on move-selection exploration, while in RQ2 we
evaluate VAM with engine-play position generation and compare against baselines under the same prompt interface and engine verifier.
RQ3 directly contrasts engine-play versus fixed-dataset training for VAM.

\paragraph{Puzzle benchmark}
We evaluate one-step move selection on the Lichess puzzle positions in Searchless
Chess~\cite{ruoss2024grandmaster}.
Each position provides a FEN string, an ordered UCI legal-move list, and a provided puzzle solution move.
We use all 10{,}000 positions as a held-out test set for puzzle evaluation.
We treat the provided solution move as the target action for puzzle accuracy.

\paragraph{Full-game benchmark}
We evaluate end-to-end play by playing complete games from the standard initial position against a fixed Stockfish
opponent (Skill Level 0), reported separately for depth 1 and depth 5.
Appendix~\ref{sec:appendix_experiments_games} specifies game termination rules, invalid-move handling, and ACPL
computation details.

\paragraph{Metrics}
For puzzles, we report pass@1 accuracy for selecting the provided puzzle solution under the
candidate set.
For full games, we report ACPL computed by a separate Stockfish analyzer that runs at depth 20, serving as an expert
evaluator.
Appendix~\ref{sec:appendix_experiments_games} specifies the exact ACPL computation and aggregation used by the evaluation
harness.

\subsection{Baselines}
\label{sec:experiments_baselines}
We compare VAM against GRPO baselines that use the same engine verifier and strict UCI output contract but omit
iterative pruning.
This setting follows Chess-R1~\citep{hwang2025can}, with the modification that we use strict UCI (instead of SAN) to
avoid SAN annotations such as check or checkmate acting as additional tactical cues.
To probe how reward design affects exploration under GRPO, we evaluate three reward definitions derived from the engine
signal: expected score, win rate, and ranking among moves.
We include a rejection-sampled supervised fine-tuning (SFT) baseline trained on the same fixed dataset as a
supervised reference point.
Appendix~\ref{sec:appendix_experiments_details} provides additional training details.

\paragraph{Reward definitions from the engine signal.}
For each position $s$ and legal move $a \in \mathcal{A}(s)$, Stockfish~\citep{stockfish} provides an evaluation that we convert into a scalar verifier score
$\mu(s,a)\in[0,1]$.
In our main experiments we use a win/draw/loss (WDL) based expected-score proxy:
if the engine reports win/draw/loss likelihoods $\big(p_W(s,a),p_D(s,a),p_L(s,a)\big)$ (with $p_W+p_D+p_L=1$),
we define
\begin{equation}
\mu_{\text{exp}}(s,a) \;=\; p_W(s,a) + \tfrac{1}{2}p_D(s,a).
\label{eq:mu-exp}
\end{equation}
We consider two additional reward variants for the GRPO baselines:
\emph{(i) win-rate reward} $\mu_{\text{win}}(s,a)=p_W(s,a)$, and
\emph{(ii) rank-based reward} computed from the within-position ordering of $\mu_{\text{exp}}(s,a)$:
\begin{equation}
\mu_{\text{rank}}(s,a) \;=\; 1 - \frac{\mathrm{rank}_{\downarrow}\!\big(\mu_{\text{exp}}(s,a)\big)-1}{|\mathcal{A}(s)|-1},
\label{eq:mu-rank}
\end{equation}
where $\mathrm{rank}_{\downarrow}$ assigns rank $1$ to the highest-scoring move (ties broken deterministically).
Unless stated otherwise, VAM and GRPO baselines use $\mu_{\text{exp}}$.

\subsection{Results}
\label{sec:experiments_results}
\begin{figure*}[t]
    \centering
    \includegraphics[width=0.7\linewidth]{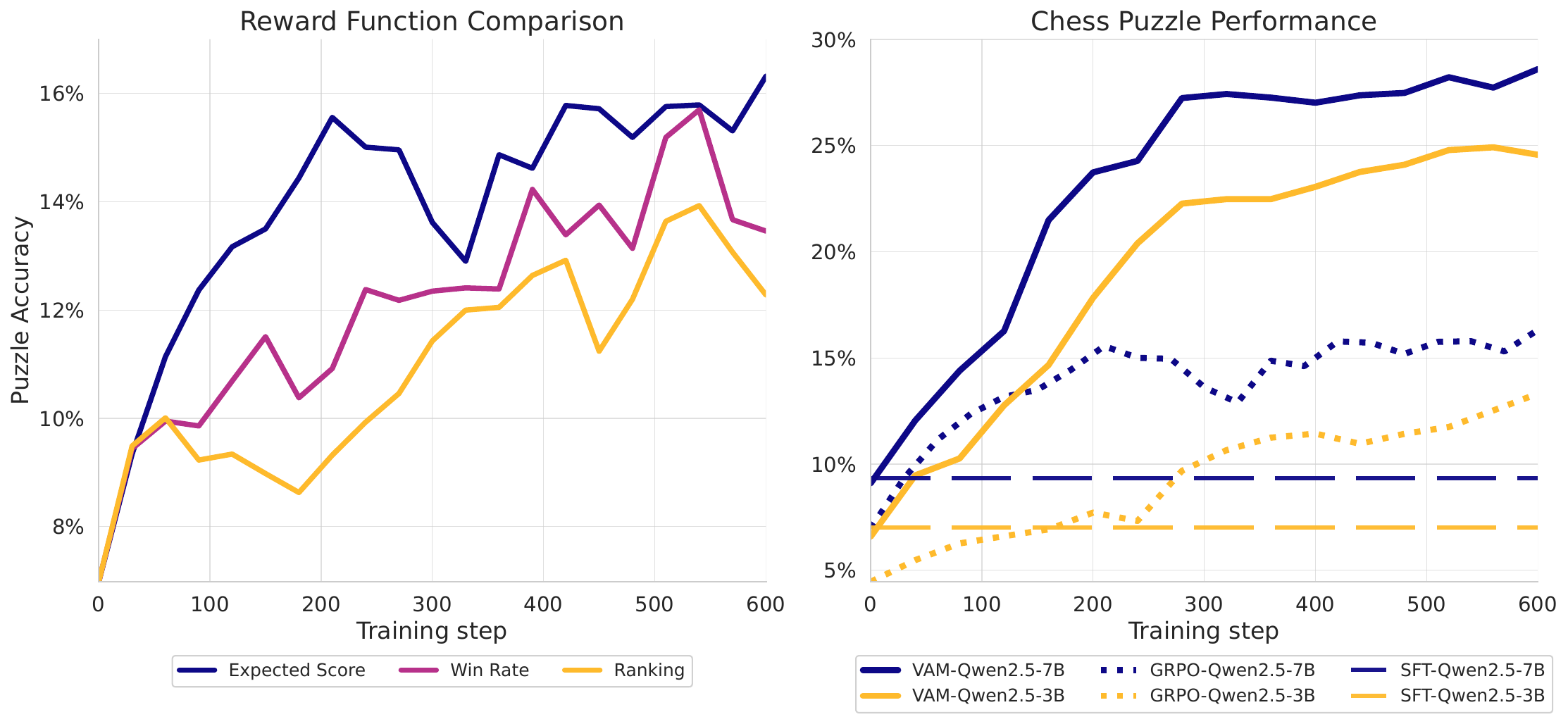}
\caption{
\textbf{Performance on held-out chess puzzles.}
Pass@1 accuracy for selecting the dataset-provided solution move on the 10{,}000-position test set.
The left panel compares fixed-dataset GRPO baselines under three engine-derived reward signals (expected-score proxy
$\mu_{\text{exp}}$, win-rate $\mu_{\text{win}}$, and rank-based $\mu_{\text{rank}}$; Eqs.~\ref{eq:mu-exp} and \ref{eq:mu-rank}).
The right panel compares VAM with iterative action-space pruning against GRPO under the same fixed dataset and matched
rollout budget for Qwen2.5-3B-Instruct and Qwen2.5-7B-Instruct.
The horizontal axis counts GRPO gradient updates, and each point evaluates the current policy with fixed decoding
settings over the full test set.
Dashed horizontal lines indicate rejection-sampled SFT baselines.
}
    \label{fig:rq1_puzzle_acc}
    \vspace{-0.5cm}
\end{figure*}

\paragraph{RQ1: exploration on puzzles}

Figure~\ref{fig:rq1_puzzle_acc} studies learning dynamics on the move-selection puzzle task.
In the left panel, we vary only the reward definition used by GRPO and observe clear differences in learning curves,
with the expected-score signal producing the strongest curve among the reward variants we tested.
For consistency in subsequent comparisons, we use expected score as the engine-derived reward for GRPO baselines and VAM.
In the right panel, we compare VAM against GRPO under the same fixed dataset and find that VAM reaches higher
puzzle accuracy for both model sizes.
Both GRPO and VAM outperform the rejection-sampled SFT baseline on held-out puzzles, suggesting that the SFT setup does
not generalize well to unseen puzzle positions.
Interestingly, the smaller Qwen2.5-3B-Instruct model trained with VAM surpasses the larger Qwen2.5-7B-Instruct model
trained with GRPO.

\paragraph{RQ2: full-game play}

\begin{figure*}[t]
    \centering
    \includegraphics[width=0.8\linewidth]{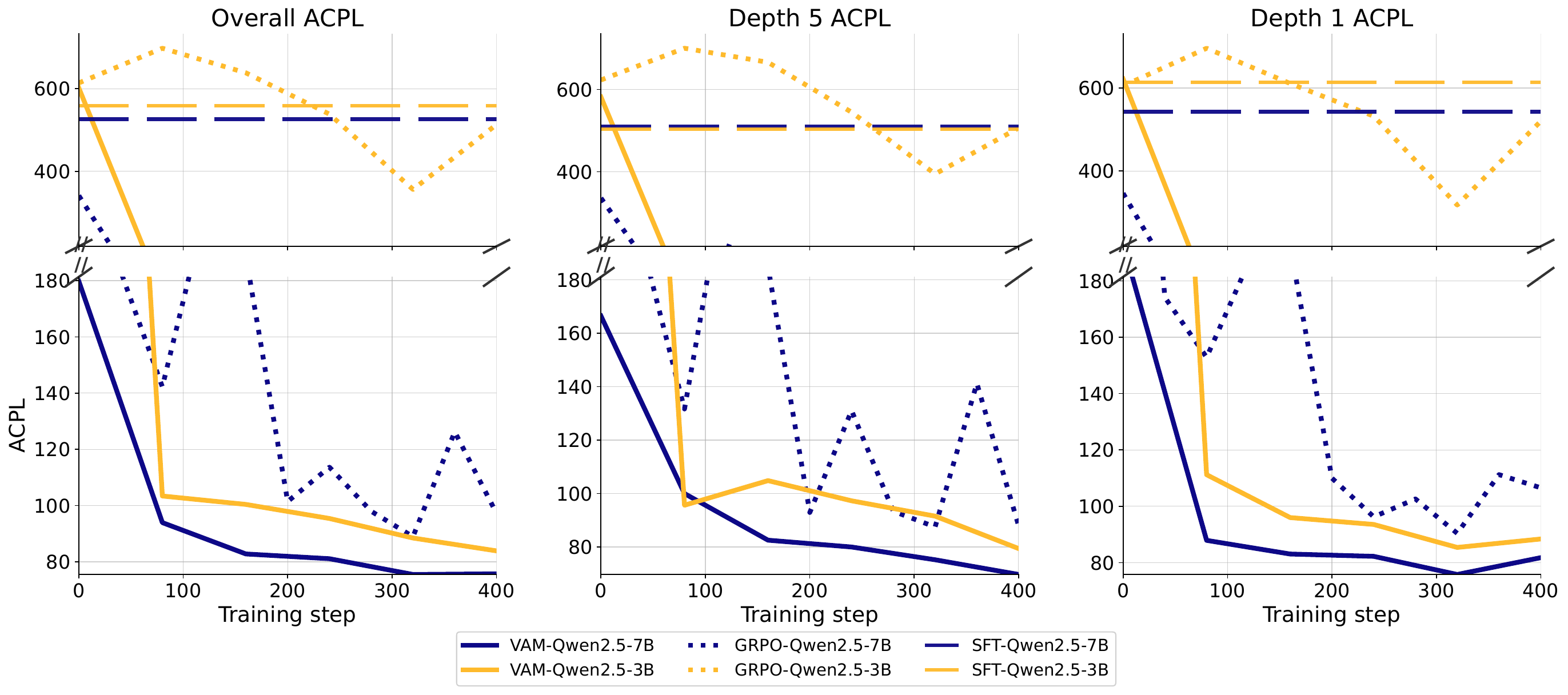}
    \caption{
    \textbf{Move quality in full games against Stockfish.}
    Average centipawn loss (ACPL; lower is better) during evaluation against fixed Stockfish opponents at depth 1 and
    depth 5, shown for Qwen2.5-3B-Instruct and Qwen2.5-7B-Instruct.
    We report overall ACPL as well as per-opponent ACPL trajectories over training.
    VAM is trained with engine-play position generation and iterative action-space pruning, while the GRPO baseline uses the
    same verifier and prompt interface but omits pruning.
    Dashed horizontal lines indicate rejection-sampled SFT baselines.
    VAM consistently reaches lower ACPL across opponents and model sizes, indicating stronger move quality under
    full-game play.
    }
    \label{fig:full_game_acpl}
    \vspace{-0.2cm}
\end{figure*}

Figure~\ref{fig:full_game_acpl} reports ACPL during full-game evaluation over the course of training.
Across both evaluation opponents (depth 1 and depth 5), VAM achieves lower ACPL than the GRPO baselines, indicating
stronger move quality in complete games.
The rejection-sampled SFT baseline yields substantially higher ACPL, suggesting that the SFT setup does not generalize
to full-game play under our evaluation protocol.
Among RL-based methods, iterative pruning provides a further improvement over standard GRPO.
Because ACPL becomes harder to reduce as it gets smaller, persistent gaps at lower ACPL correspond to meaningful
differences in play.

\paragraph{RQ3: engine-play versus fixed-dataset regimes}

\begin{figure*}[t]
    \centering
    \includegraphics[width=0.85\linewidth]{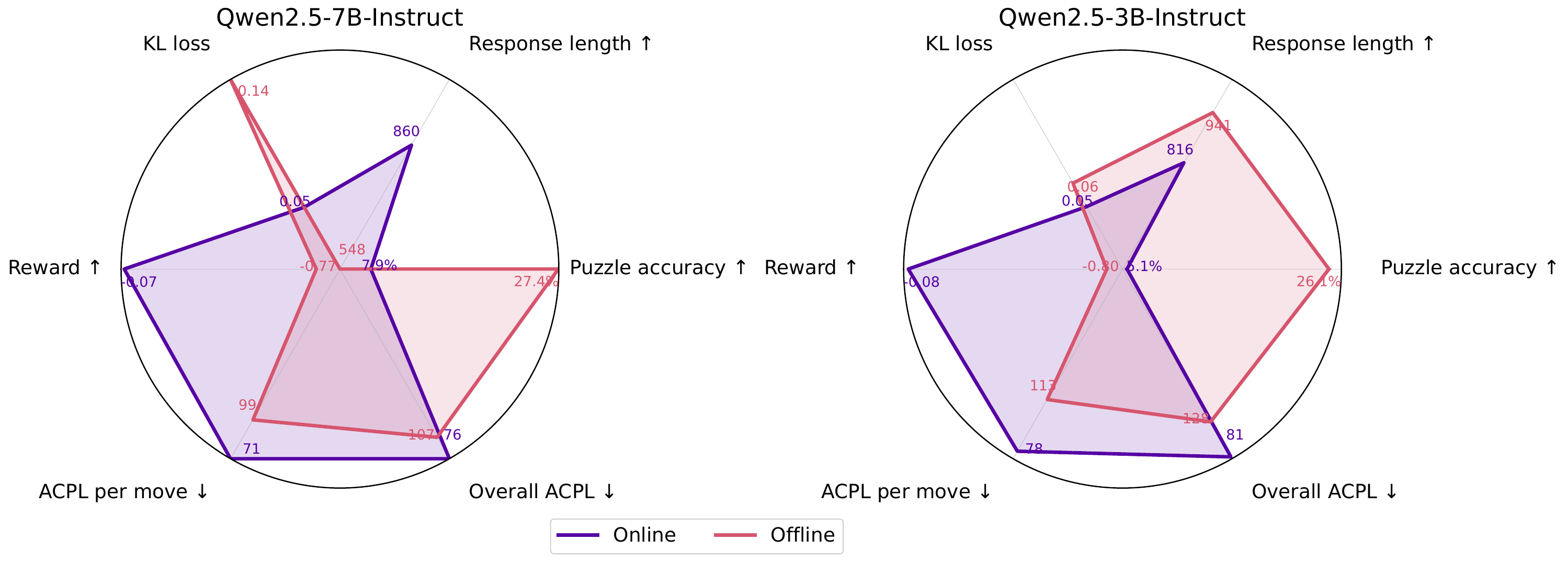}
\caption{
    \textbf{Fixed-dataset versus engine-play training regimes for VAM.}
    Radar plots summarize performance and training statistics for VAM trained either on a fixed dataset or with
    engine-play position generation via games against a fixed engine opponent, shown for Qwen2.5-7B-Instruct (left) and Qwen2.5-3B-Instruct (right).
    Axes report pass@1 puzzle accuracy, KL loss to the reference policy, average response length, mean verifier reward,
    and full-game move quality measured by ACPL (overall and per-move).
    Within each model, each metric is linearly normalized across the two regimes for visualization, and numeric
    annotations indicate the raw values.
    The profiles highlight a trade-off between one-step puzzle performance and full-game play quality across regimes.
    }
    \label{fig:rq3_online_offline_radar}
\end{figure*}

Figure~\ref{fig:rq3_online_offline_radar} contrasts VAM trained on a provided fixed dataset with VAM trained on states
generated via games against a fixed engine opponent.
We observe a clear trade-off between one-step selection and long-horizon play: fixed-dataset training yields stronger puzzle
accuracy, while engine-play position generation yields better full-game ACPL.
Beyond these headline metrics, the engine-play regime is associated with longer responses and smaller KL loss, which is
consistent with training on a state distribution induced by the current policy.

\section{Discussion}
\label{sec:discussion}

\paragraph{When VAM is a good fit}
VAM is most appropriate when the environment can enumerate a finite action set per state and when membership in the
admissible set can be checked deterministically.
In such settings, mask compliance becomes a reliable signal that can be used for training (as a hard penalty) and for
evaluation (as a format and legality metric), while preserving on-policy sampling because we do not apply logit-level
masking or post-hoc filtering.

\paragraph{Exploration as a controllable interface choice}
Iterative action-space pruning turns repeated sampling into a controlled exploration procedure: across rounds, the model
is forced to propose different actions because previously sampled valid actions are removed from the prompt-specified
candidate set.
This perspective helps disentangle two sources of failure in RL post-training, limited policy entropy and limited
action-space coverage under repeated sampling.

\section{Conclusion}
\label{sec:conclusion}

We introduced Verbalized Action Masking (VAM), a prompt-level action-space interface for RL post-training with verifiable
rewards, along with an iterative action-space pruning procedure that improves exploration by preventing repeated sampling
of the same actions within a state.
We instantiated these ideas in text-only chess move selection with strict UCI outputs and evaluated them on both one-step
puzzle selection and full-game play against an engine opponent.
Empirically, we found that expected score is the strongest reward choice for GRPO baselines on puzzles, and that VAM
substantially improves puzzle accuracy and full-game ACPL over those baselines across model sizes.
We also observed complementary strengths between fixed-dataset and engine-play regimes, suggesting that state distribution is a
key design choice for RL post-training in structured action spaces.

\textbf{Limitations and future work.} Chess is an ideal testbed because legal moves are enumerable and both action
validity (strict parsing and move legality) and move quality can be checked deterministically with an engine. Many
structured decision problems do not offer these properties, so applying VAM in domains with very large action spaces or
without strong deterministic evaluators will require dynamic candidate generation and approximate verifiers.

\section*{Acknowledgements}

This work was supported in part by NSF grant IIS-2046640 (CAREER). This work was supported by the Engineering and Physical Sciences Research Council [grant number EP/Y003187/1 and UKRI849].

\section*{Impact Statement}

This paper presents work whose goal is to advance the field of machine learning, specifically methods for reinforcement
learning post-training of large language models under verifiable constraints.
There are many potential societal consequences of improved LLM capabilities and of broader adoption of RL post-training,
none of which we feel must be specifically highlighted here.

\bibliography{ref}
\bibliographystyle{icml/icml2026}

\clearpage
\appendix
\section{Additional experimental details}
\label{sec:appendix_experiments_details}

This appendix specifies evaluation details that are important for reproducibility, but are too low-level to include in
the main experimental narrative.

\subsection{Training data}
\label{sec:appendix_experiments_training}

\paragraph{Training position sources}
We consider two sources of training positions.
First, a fixed dataset aligned with Chess-R1~\cite{hwang2025can}, containing 100{,}000 chess positions.
Second, an on-policy dataset generated via engine-play against an engine opponent, where we record encountered game
states (from both sides) and query the engine to obtain $\mu(s,a)$ over the legal move list (Algorithm~\ref{alg:online_play}).

\paragraph{Fields per training position}
Each training example includes the position (FEN), an ordered legal-move list in strict UCI, and an engine-provided
per-move value map $\mu(s,a)$ used for reward computation.
For the fixed dataset, each example also includes a dataset-provided solution move in UCI.
For VAM-style selection prompts, we additionally specify an ordered candidate subset \texttt{allowed\_moves} that serves
as the initial mask $\mathcal{M}_0$; when a candidate subset is not provided by the dataset, $\mathcal{M}_0$ can be set
to the full legal set $\mathcal{A}(s)$ (\S\ref{sec:methods_reward}).

\subsection{Rejection-sampled supervised fine-tuning (SFT) baseline}
\label{sec:appendix_experiments_sft}

We construct the supervised baseline dataset from the same fixed 100{,}000-position dataset described in
\S\ref{sec:appendix_experiments_training}.
Each example provides a dataset-provided solution move in UCI.
We use the baseline prompt interface without a candidate subset and apply rejection sampling over model generations,
keeping only generations whose parsed strict UCI move matches the dataset-provided solution (rejecting malformed or
illegal outputs).
This procedure yields 45{,}806 accepted examples for Qwen2.5-3B-Instruct (5.73\% acceptance) and 71{,}181 accepted
examples for Qwen2.5-7B-Instruct (8.90\% acceptance).
Table~\ref{tab:appendix_experiments_hyperparams} lists the settings used for this baseline.

\subsection{Engine-derived verifier scores}
\label{sec:appendix_experiments_verifier}

\paragraph{Stockfish-based scoring}
We instantiate the verifier score $\mu(s,a)$ using Stockfish~\cite{stockfish}.
In the main experiments, we use the expected-score proxy $\mu_{\text{exp}}$ (Eq.~\ref{eq:mu-exp}) derived from the
engine's win/draw/loss probabilities, and define the target action within a candidate set as the $\mu$-best move with a
deterministic tie-break (\S\ref{sec:methods_reward}).
\Cref{sec:experiments_baselines} defines additional reward variants ($\mu_{\text{win}}$ and $\mu_{\text{rank}}$) used in
the GRPO reward-function comparison.

\paragraph{Reward assignment under the candidate mask}
For both VAM and GRPO baselines, the reward is computed from the decoded move and the engine metadata for the current
position.
If the decoded move is well-formed, legal, and (for selection prompts) inside \texttt{allowed\_moves}, it receives the
engine-derived score $\mu(s,a)$.
Otherwise (missing tag, malformed UCI, illegal move, or out-of-candidate move), we treat the output as invalid and
assign a fixed negative penalty, terminating the rollout (\S\ref{sec:methods}).

\subsection{Hyperparameters}
\label{sec:appendix_experiments_hyperparams}

Table~\ref{tab:appendix_experiments_hyperparams} summarizes the main hyperparameters used across experiments.
Shared optimization settings are matched across RL methods for fair comparisons, and VAM-specific controls are listed in
the VAM column.
The last two columns report the rejection-sampled SFT baseline settings.

\begin{table*}[!tbp]
\centering
\caption{Hyperparameters for GRPO, VAM, and SFT.}
\label{tab:appendix_experiments_hyperparams}
\setlength{\dashlinedash}{1.0pt}
\setlength{\dashlinegap}{1.2pt}
\renewcommand{\arraystretch}{1.06}
\begin{tabular*}{\textwidth}{@{\extracolsep{\fill}}lcc:ll@{}}
\toprule
Parameter & GRPO & \multicolumn{1}{c}{VAM} & Parameter & SFT \\
\midrule
Batch size & 128 & 128 & Global batch size & 32 \\
Mini batch size & 128 & 128 & Total source prompts & 100{,}000 \\
GRPO group size & 8 & 8 & Samples per prompt & 8 \\
Sampling temperature & 1.0 & 1.0 & Sampling temperature & 0.6 \\
Sampling top-$p$ & 1.0 & 1.0 & Sampling top-$p$ & 0.9 \\
Max response tokens & 2000 & 2000 & Max response tokens & 2000 \\
Epochs & 1 & 1 & Epochs & 2 \\
Optimizer & AdamW & AdamW & Optimizer & AdamW \\
Learning rate & 1e-6 & 1e-6 & Base learning rate & 1e-5 \\
LR schedule & constant & constant & LR schedule & cosine \\
Warmup ratio & 10 steps & 10 steps & Warmup ratio & 0.1 \\
Weight decay & 0.01 & 0.01 & Weight decay & 0.01 \\
Gradient clipping & 1 & 1 & Gradient clipping & 1 \\
KL loss coefficient & 1e-3 & 1e-3 & \#SFT examples & Qwen2.5-3B: 45{,}806 \\
\cdashline{5-5}
Entropy coefficient & 0 & 0 &  & Qwen2.5-7B: 71{,}181 \\
Clip ratio & 0.2 & 0.2 &  &  \\
Reward function & $\mu_{\text{exp}}/\mu_{\text{win}}/\mu_{\text{rank}}$ & $\mu_{\text{exp}}$ &  &  \\
$R_{\max}$ & $\diagup$ & 4 &  &  \\
Online opponent & $\diagup$ & stockfish (skill 0, depth 1) &  &  \\
\bottomrule
\end{tabular*}
\end{table*}

\subsection{Prompt interface and parsing}
\label{sec:appendix_experiments_parsing}

\paragraph{Inputs per decision}
Each decision includes the FEN string, an explicit side-to-move field, and an ordered list of legal moves in strict
lowercase UCI.
In the candidate-restricted (selection) setting we additionally provide an ordered subset \texttt{allowed\_moves}
(also called \texttt{considered\_moves\_uci}) and require the model to choose from that subset.

\paragraph{Required output}
The model must output a single move inside a \texttt{\textless uci\_move\textgreater} tag, with a strict lowercase UCI
string as payload (for example, \texttt{e2e4}).
Promotions include the piece letter (for example, \texttt{e7e8q}).
Castling is represented in UCI as the king move (for example, \texttt{e1g1} for white kingside castling).

\paragraph{Validity checks}
We treat any unparsable output, illegal move, or (for selection prompts) out-of-candidate move as invalid.
For selection prompts, the decoded move must match one of the provided candidate strings exactly.

\subsection{Puzzle evaluation details}
\label{sec:appendix_experiments_puzzles}

\paragraph{Data fields}
Each puzzle row provides \texttt{fen}, an ordered \texttt{legal\_moves\_uci} list, and a provided solution move in UCI.
Some rows additionally include per-move value maps defined over the legal move list, stored as JSON keyed by UCI
strings.
The value maps include an engine-derived expected score or win-probability proxy in $[0,1]$.

\paragraph{Targets and metrics}
We treat the dataset-provided solution move as the target action $a^\star$ for evaluation.
The main paper reports pass@1 accuracy (success if $\hat a = a^\star$), where invalid outputs count as failures.
We additionally track format and legality compliance and, when value maps are available, the mean selected-move value
$\mathbb{E}[\mu(s,\hat a)]$.

\subsection{Full-game evaluation details}
\label{sec:appendix_experiments_games}

\paragraph{Match setup}
Games start from the standard initial position and are capped at 200 plies.
We do not automatically claim draws.
Unless stated otherwise, we play 50 games per evaluation depth (depth 1 and depth 5), with per-depth color balancing
and a deterministic shuffle.

\paragraph{Invalid moves}
If the model output is unparsable or illegal, we retry up to three attempts.
Persistent failure is treated as resignation or forfeit, which affects win rate and can yield worst-case ACPL when no
valid moves are played.

\paragraph{ACPL computation}
We compute centipawn loss using a separate Stockfish analyzer at depth 20, with approximately 1 second of analysis per
move.
Mate scores are mapped to 1000, and centipawn values are clamped to a fixed cap, matching the evaluation harness
semantics.
Per-move centipawn loss (CPL) uses the before versus after evaluations, with evaluations converted to the mover's point
of view and clipped at 0:
\begin{equation}
\mathrm{CPL} \;=\; \max\!\bigl(0, E_{\text{before}} - E_{\text{after}}\bigr).
\end{equation}
We first compute per-game ACPL as the mean CPL over the model's moves within that game.
We then report two aggregates across a batch of games: (i) \emph{overall ACPL}, the mean of per-game ACPL values (each
game equally weighted), and (ii) \emph{ACPL per move}, a move-weighted mean over all model moves in the batch.

\subsection{Prompt}
\label{sec:appendix_experiments_prompt}
\noindent We include the exact prompt templates used in our experiments in \Cref{fig:appendix_prompt_vam,fig:appendix_prompt_baseline}.
\begin{figure*}[t]
\centering
\begin{minipage}{\textwidth}
\begin{tcolorbox}[
    enhanced,
    colback=promptBg,
    colframe=promptFrame,
    arc=1.0mm,
    boxrule=0.35pt,
    boxsep=0.4mm,
    left=3.2mm,
    right=3.2mm,
    top=2.0mm,
    bottom=0.35mm,
    left skip=0mm,
    right skip=0mm,
    borderline west={1.0pt}{0pt}{promptAccent},
    title={Prompt used for VAM},
    fonttitle=\bfseries\promptfont\footnotesize,
    coltitle=white,
    attach boxed title to top left={xshift=1.5mm, yshift=-1.1mm},
    boxed title style={
        colback=promptAccent,
        colframe=promptAccent,
        arc=0.9mm,
        boxrule=0pt,
        top=0.35mm, bottom=0.35mm, left=0.95mm, right=0.95mm
    },
    before upper=\setlength{\parindent}{0pt}\setlength{\parskip}{0.2em}\setlength{\emergencystretch}{1.5em}\footnotesize\ttfamily\color{promptText}\sloppy
]
\noindent You are a helpful assistant and a professional chess player.

You will be given: \\
- a FEN position, \\
- the full list of legal moves, \\
- and a smaller list called allowed\_moves.

allowed\_moves is a preselected shortlist intended to concentrate attention on the most relevant, often-promising candidates in this position.
It is an unordered set: the order of moves in allowed\_moves is arbitrary and does not indicate strength or preference.
Your job is to analyze the position deeply and choose the single best move from allowed\_moves.

Output format (required): \\
- Put your reasoning inside <think>...</think>. \\
- Put your final chosen move inside <uci\_move>...</uci\_move>. \\
- Output exactly one move in <uci\_move> and nothing else outside these tags. \\
- Output exactly: <think>...</think><uci\_move>...</uci\_move> and stop after </uci\_move>.

How to think (guidance): \\
- Evaluate the position (material, king safety, piece activity, pawn structure, threats, initiative). \\
- Check for forcing tactics (checks, captures, direct threats) for both sides. \\
- Use allowed\_moves as the candidate set; compare the key candidates from that set. \\
- For each key candidate, consider the opponent's most likely/best reply and the resulting position. \\
- Choose the move in allowed\_moves that yields the best expected outcome for the side to move.

Selection constraint (must follow): \\
- The content of <uci\_move> MUST be a valid UCI move string AND MUST match an element of allowed\_moves exactly. \\
- Do not invent moves, and do not output resign/None/N/A.

UCI notation reminder: \\
- Use from-square + to-square, e.g. e2e4, g1f3, a7a8q (promotion uses the piece letter suffix).

Reminder of chess rules: \\
- Bishops move diagonally. \\
- Rooks move horizontally or vertically. \\
- Knights jump in an L-shape. \\
- Queens combine rook and bishop movement. \\
- Kings move one square in any direction. \\
- Pawns move forward, capture diagonally, and can promote.

Position (FEN): \{\{ FEN \}\} \\
Legal moves (UCI): \{\{ legal\_moves\_uci\_list | join(', ') \}\} \\
Allowed moves (UCI): \{\{ considered\_moves\_uci\_list | join(', ') \}\}

Think deeply, then output the single best allowed move.
\end{tcolorbox}
\end{minipage}
\caption{Prompt used for Verbalized Action Masking (VAM).}
\label{fig:appendix_prompt_vam}
\end{figure*}

\begin{figure*}[t]
\centering
\begin{minipage}{\textwidth}
\begin{tcolorbox}[
    enhanced,
    colback=promptBg,
    colframe=promptFrame,
    arc=1.0mm,
    boxrule=0.35pt,
    boxsep=0.4mm,
    left=3.2mm,
    right=3.2mm,
    top=2.0mm,
    bottom=0.35mm,
    left skip=0mm,
    right skip=0mm,
    borderline west={1.0pt}{0pt}{promptAccent},
    title={Prompt used for baselines},
    fonttitle=\bfseries\promptfont\footnotesize,
    coltitle=white,
    attach boxed title to top left={xshift=1.5mm, yshift=-1.1mm},
    boxed title style={
        colback=promptAccent,
        colframe=promptAccent,
        arc=0.9mm,
        boxrule=0pt,
        top=0.35mm, bottom=0.35mm, left=0.95mm, right=0.95mm
    },
    before upper=\setlength{\parindent}{0pt}\setlength{\parskip}{0.2em}\setlength{\emergencystretch}{1.5em}\footnotesize\ttfamily\color{promptText}\sloppy
]
\noindent You are a helpful assistant who plays chess professionally.
The assistant first thinks through the reasoning process internally and then provides the user with the best move.
The reasoning process and the answer must be enclosed within <think> </think> and <uci\_move> </uci\_move> tags, respectively.
The reasoning process should describe how you analyze the position and decide on the best move, including: \\
- A strategic evaluation of the position. \\
- A comparison of key candidate moves. \\
- For each candidate, consider the opponent's likely response and outcome. \\
- Conclude with a clear justification for your final choice.
The answer must be in UCI notation, using the from-square and to-square (e.g., e2e4, g1f3, a7a8q).
Now, the user provides a FEN string, and a list of legal moves for the given board.
After analyzing the position, clearly state the best move in UCI notation within <uci\_move> </uci\_move> tags. i.e., <uci\_move> e2e4 </uci\_move>

Reminder of chess rules: \\
- Bishops move diagonally. \\
- Rooks move horizontally or vertically. \\
- Knights jump in an L-shape. \\
- Queens combine rook and bishop movement. \\
- Kings move one square in any direction. \\
- Pawns move forward, capture diagonally, and can promote.

Current FEN string: \{\{ FEN \}\} \\
Legal moves (UCI): \{\{ legal\_moves\_uci\_list | join(', ') \}\}

Let's think step by step.
\end{tcolorbox}
\end{minipage}
\caption{Prompt used for baseline methods.}
\label{fig:appendix_prompt_baseline}
\end{figure*}

\end{document}